\documentclass[runningheads,a4paper]{llncs}

\usepackage{amssymb}
\usepackage{graphicx}
\usepackage{epstopdf}
\usepackage{amsmath}
\usepackage{float}
\usepackage{booktabs}
\usepackage{multirow}
\usepackage{multicol}

\usepackage{url}
\urldef{\mailsa}\path|s.imangaliyev@vumc.nl|

\newcommand{\keywords}[1]{\par\addvspace\baselineskip
\noindent\keywordname\enspace\ignorespaces#1}

\begin{document}

\mainmatter  

\title{Classification of Quantitative Light-Induced Fluorescence Images Using Convolutional Neural Network}

\titlerunning{Classification of QLF-images Using Convolutional Neural Network}

%
%
\author{Sultan Imangaliyev\inst{1,2,6}
\thanks{Corresponding author.}\and
Monique H. van der Veen\inst{3} \and Catherine M. C. Volgenant\inst{3} \and Bruno G. Loos\inst{3} \and Bart J. F. Keijser\inst{3,4} \and Wim Crielaard\inst{3} \and Evgeni Levin\inst{5,6}}

\authorrunning{Imangaliyev et al.}
\institute{VU University Medical Center Amsterdam, Amsterdam, The Netherlands,
\and
Cancer Center Amsterdam, Amsterdam, The Netherlands,
\and
Academic Centre for Dentistry Amsterdam, Amsterdam, The Netherlands,
\and
Netherlands Organisation for Applied Scientific Research, Zeist, The Netherlands,
\and
Academic Medical Center, Amsterdam, The Netherlands
\and
Horaizon BV, Rotterdam, The Netherlands\\
\mailsa}
%
%

\toctitle{Classification of Quantitative Light-induced Fluorescence Images Using Convolutional Neural Network}
\tocauthor{S. Imangaliyev et al.}
\maketitle

\begin{abstract}
Images are an important data source for diagnosis and treatment of oral diseases. The manual classification of images may lead to misdiagnosis or mistreatment due to subjective errors. In this paper an image classification model based on Convolutional Neural Network is applied to Quantitative Light-induced Fluorescence images. The deep neural network outperforms other state of the art shallow classification models in predicting labels derived from three different dental plaque assessment scores. The model directly benefits from multi-channel representation of the images resulting in improved performance when, besides the Red colour channel, additional Green and Blue colour channels are used.
\end{abstract}

\section{Introduction}
Diagnosis and therapy in many areas of medicine, including dentistry, nowadays extensively rely on technological advances in biomedical imaging. One of the challenges in the diagnosis of dental patients during daily practice is assessment of their dental plaque level. A novel way to look at this plaque is the use of a Quantitative Light-induced Fluorescence (QLF) camera. When the QLF-camera is used some dental plaque fluoresces red, which is suggested to be an indication for the pathogenicity of the dental plaque \cite{van2016dynamics}.

In this paper we apply deep artificial neural network on QLF-images to make a predictive classification model, where class separation is based on the amount of red fluorescent dental plaque disclosed in such images. Although both intra-examiner and inter-examiner reliability of manual assessment of QLF-images are shown to be high \cite{volgenant2016comparison}, this may become expensive and laborious if the number of images is large. Therefore, there is a need to automate this procedure by implementing a computer-based system for assessment of QLF-images. Existing computer programs developed for this goal have several drawbacks which limit efficiency of QLF-images assessment. They require that the images must have been captured under the fixed circumstances such as camera geometry, focal distance and ambient light conditions \cite{kim2014monitoring}, which is hard to achieve under clinical settings.

The problem mentioned above could be solved by the use of Deep Learning models, because descriptive features can be learnt directly from raw data representations \cite{LeCun2015} being insensitive to ambient conditions and natural image variability. Since images have a special two-dimensional structure, a group of Deep Learning methods called Convolutional Neural Network (CNN) explicitly uses the advantages of such a representation \cite{jarrett2009best,lecun2010convolutional}. Applications of CNN may include both non-biological \cite{david2016deeppainter} and biological images \cite{esteva2017dermatologist}.

The aim of this paper is to describe the novel application of CNN to QLF-images obtained during clinical intervention study \cite{van2016dynamics}. Furthermore, we compare the performances of the CNN and several state of the art classification models. We tested all of these models on three existing plaque assessment scoring systems. We also checked the influence of adding various colour channels on the model performance. Possible differences were explained based on the biological nature of the problem and based on the properties of these models. Previous studies on this topic either focused on only a single plaque scoring system without providing detailed analysis of results \cite{imangaliyev2016deep} or used small dataset of different images and different network architecture \cite{kang2006dental}.

\section{Materials and Methods}
\subsection{Convolutional Neural Networks}
Many of the modern deep learning models utilize very deep architectures to achieve superhuman performance in solving object recognition problems \cite{simonyan2014very,szegedy2015going}. One of such architectures is a novel ultra-deep residual learning network (ResNet) \cite{he2016deep}. This architecture can be implemented by adding so called 'shortcut connections' \cite{he2016identity} which skip one or more layers. They perform a mapping so that their outputs are added to the outputs of the stacked layes. The whole network can be trained and implemented by using common libraries without modifying the solvers, hence adding neither extra parameters nor computational complexity. ResNet and many other architectures \cite{jarrett2009best,lecun2010convolutional} use convolutional operator in extracting useful feature mappings in image classification task. Generally, given the filter $K \in \mathbb{R}^{(2h_1+1) \times (2h_2+1)}$, the discrete convolution of the image $I$ with filter $K$ is given by
\begin{align}
	\label{eq:convolution}
	\left(I \ast K\right)_{r,s} := \sum _{u = -h_1} ^{h_1} \sum _{v = -h_2}^{h_2} K_{u,v} I_{r+u,s+v}.
\end{align}

Let layer $l \in \mathbb{Z}$ be a convolutional layer. The $i^{\text{th}}$ feature map in layer $l$, denoted $Y_i^{(l)}$, is computed as
\begin{align}
	\label{eq:convolutional-layer}
	Y_i^{(l)} = B^{(l)}_{i} + \sum _{j = 1}^{m_1^{(l-1)}} K^{(l)}_{i,j} \ast Y_j^{(l-1)},
\end{align}
where $B_i^{(l)}$ is a bias matrix and $K^{(l)}_{i,j}$ is the filter of size $(2h_1^{(l)} + 1) \times (2h_2^{(l)} + 1)$ connecting the $j^{\text{th}}$ feature map in layer $(l-1)$ with the $i^{\text{th}}$ feature map in layer $l$ \cite{lecun2010convolutional}.

\subsection{Dataset}
The analyzed 427 QLF-images were taken during a clinical intervention study \cite{van2016dynamics} which was conducted at the Academic Centre for Dentistry Amsterdam. Those images were translated into a combined dataset of three colour channels with $216\times324$ raw pixel intensity values in each of them. In total, three different experiments were performed on labels derived from plaque scoring systems such as Red Fluorescent Plaque Percentage (RF-PP) \cite{van2016dynamics}, Red Fluorescent modified Quigley-Hein index (RF-mQH) \cite{volgenant2016comparison} and modified Sillness-Loe Plaque index (mSLP) \cite{weijden1993comparative}.

\subsection{Experimental Setup}
The CNN model was implemented on an NVIDIA GeForce GTX Titan X Graphics Processing Unit (GPU) using the \emph{Theano} package \cite{bergstra2011theano}. To compare the influence of different colour channels three dataset compositions were tested which are only Red, Red with Green, or full RGB representations. To compare the CNN performance with the performance of the other models, experiments were performed using various shallow classification models implemented in the \emph{Scikit-learn} package \cite{scikit-learn} such as Logistic Regression (LR), Support Vector Machines Classifier with Gaussian Kernel (SVMC-K), Support Vector Machines Classifier with Linear Kernel (SVMC-L), Gaussian Naïve Bayes Classifier (GNB), Gradient Boosting Classifier (GBC), K-Neighbors Classifier (KNC), and Random Forest Classifier (RFC).

Hyperparameters of those models were selected via an exhaustive grid search with stratified shuffled cross-validation procedure so that 80\% of the dataset was used as a training set, 10\% as a validation set, and the rest 10\% as a test set. All binary models were adapted to a multiclass setting by using a \emph{one-versus-all} approach. The predictive performance of the models was assessed by calculating the $F_1$-score \cite{sokolova2009systematic}. The reported final $F_1$-score was obtained by averaging the results of ten random shuffles with fixed test-train splits across all models.

\section{Results and Discussion}
\subsection{Model Performance Evaluation}
\begin{figure}[]
\centering
\includegraphics[height=6.0cm]{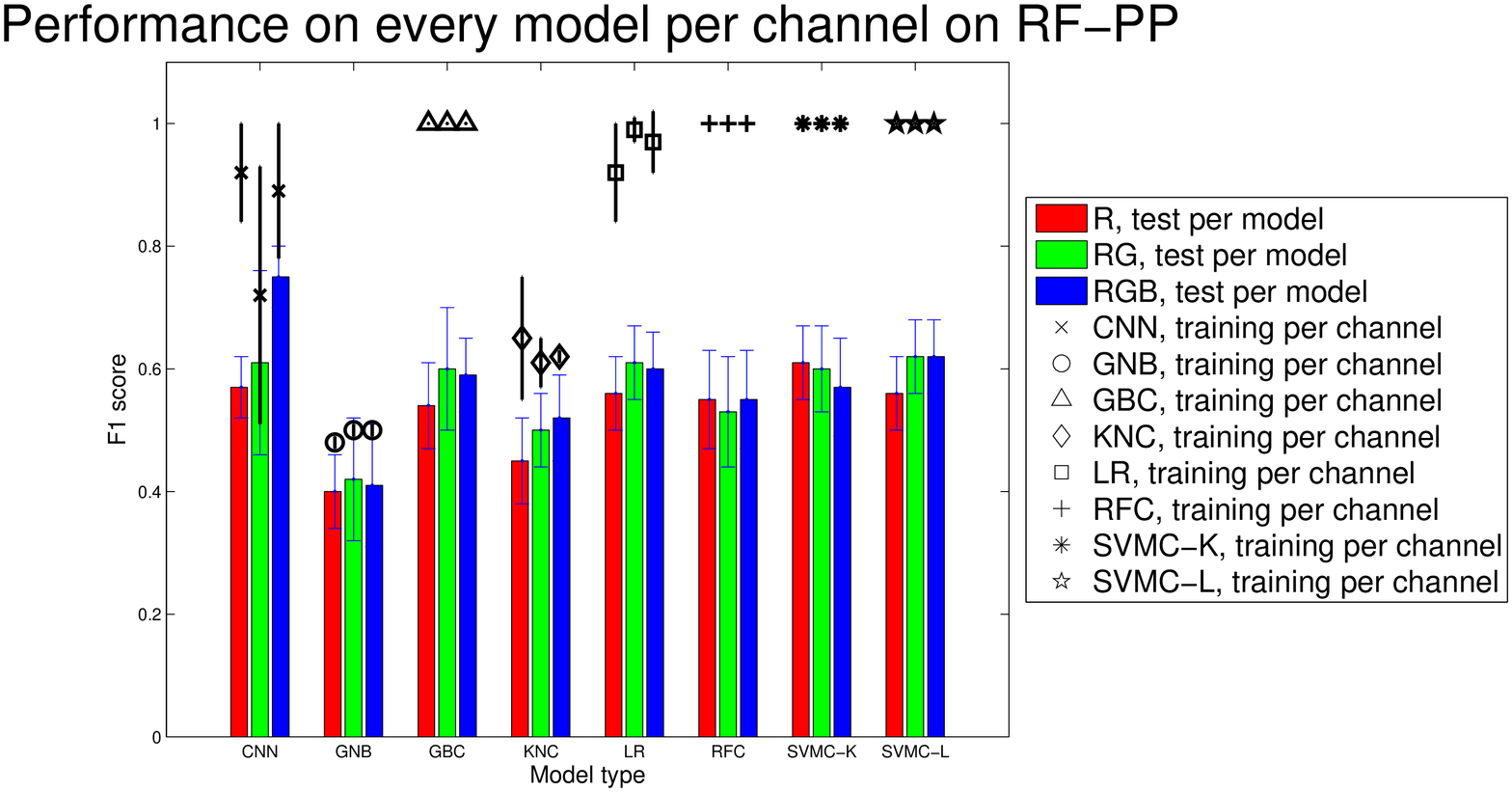}
\caption{Test and training performance of models on QLF-images using classes derived from the Red Fluorescent Plaque Percentage (RF-PP) values as labels.}
\label{fig:barchart_rfplaque}
\end{figure}

\begin{figure}[]
\centering
\includegraphics[height=6.0cm]{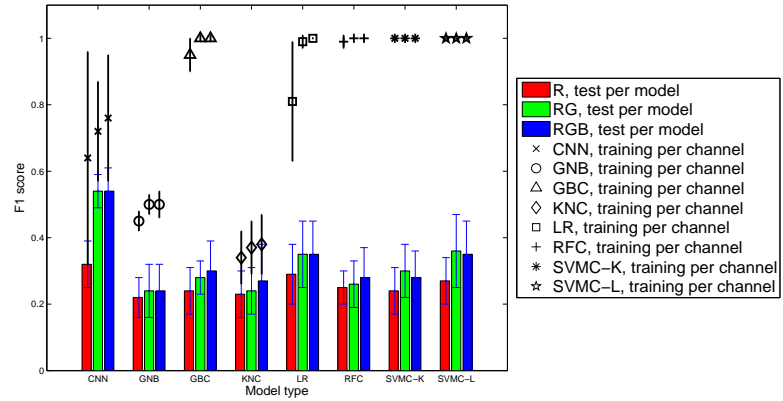}
\caption{Test and training performance of models on QLF-images using classes derived from the average Red Fluorescent modified Quigley-Hein (RF-mQH) values as labels.}
\label{fig:barchart_mqh}
\end{figure}

\begin{figure}[]
\centering
\includegraphics[height=6.0cm]{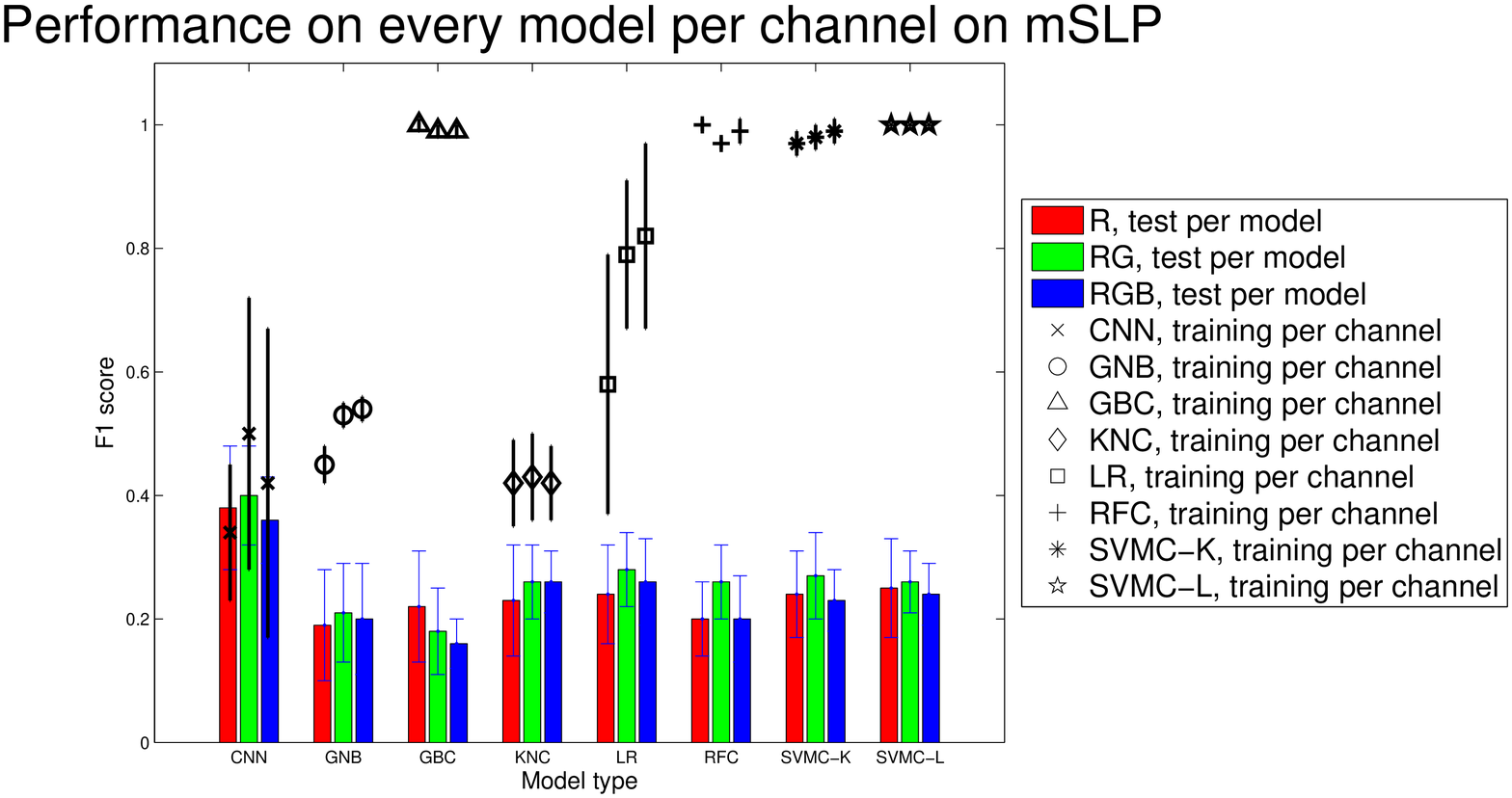}
\caption{Test and training performance of models on QLF-images using classes derived from the percentage of modified Sillness-Loe Plaque (mSLP) index values as labels.}
\label{fig:barchart_plaqueperc}
\end{figure}

Results of experiments for RF-PP, RF-mQH and mSLP labels are provided in Figure~\ref{fig:barchart_rfplaque}, Figure~\ref{fig:barchart_mqh}, and Figure~\ref{fig:barchart_plaqueperc} respectively. As it is seen from Figure~\ref{fig:barchart_rfplaque}, in the experiment with the RF-PP label, most of the models have a perfect classification performance on the training dataset, but a poor performance on the test dataset. Moreover, the results indicate that using only the Red channel results in a relatively good and comparable performance between both SVM models and Logistic Regression. Adding the Green and especially Blue channels improves the performance of CNN compared to the other models. As a result, the best model (CNN) provided a 0.76 $\pm$ 0.05 $F_1$-score on the test set and a 0.89 $\pm$ 0.11 $F_1$-score on the training set.

Similar to the experiment with RF-PP labels, results depicted in Figure~\ref{fig:barchart_mqh}, and Figure~\ref{fig:barchart_plaqueperc} clearly demonstrate the advantage of CNN over the other models, especially after adding the Green channel. As a result, the best model (CNN) provided a 0.54 $\pm$ 0.07 $F_1$-score on the test set for RF-mQH labels and a 0.40 $\pm$ 0.08 $F_1$-score on the test set for mSLP labels. However, unlike in the RF-PP case, adding the Blue channel did not improve and even decreased the performance for most of the models. Also, there is a clear difference between the performance of models applied on RF-PP and the other labels overall. Namely, even the best model's $F_1$-scores are in the interval ${[0.4, 0.55]}$ in the experiments with RF-mQH and mSLP labels, which are much less than the 0.76 achieved in experiments with the RF-PP label.

\subsection{Advantages of the Deep Learning Model}
The results of the models' predictive performance evaluation clearly demonstrated advantage of the CNN model over the other models. In general, the predictive performance of the model on previously unseen data, i.e., its generalization can be improved if certain \textit{a priori} information about the problem is added into the choice of the model architecture \cite{lecun1989backpropagation}. In case of images, domain information about the problem can be utilized by a model if such a model is able to learn spatial information between the pixels of an image. This property is explicitly embedded into the CNN model via a discrete convolution operation \cite{LeCun2015}. In the case of the QLF-images the model may learn, for example, the intensity of red colour associated with plaque, or the sharpness of edges between gingiva and teeth as well as between teeth. Classification results shown in Figures~\ref{fig:barchart_rfplaque},~\ref{fig:barchart_mqh},~\ref{fig:barchart_plaqueperc} indicate the robustness of CNN to overfitting despite image variability. Other models used in this study do not directly embed spatial information unique for image pixel representation, thus these models have poorer generalization properties and result in a lower classification performance on previously unseen data.

\begin{figure}[]
\centering
\includegraphics[height=2.5cm]{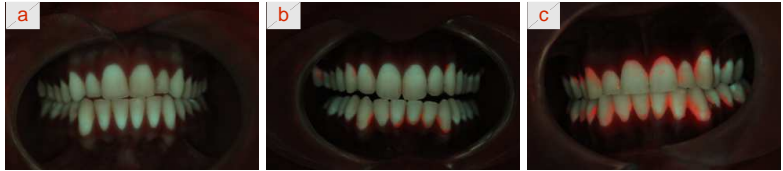}
\caption{Typical examples of QLF-images taken at the last day of the clinical intervention separated into three classes depending on different levels of plaque accumulation, for a subject with low (a), moderate (b) or high (c) red fluorescent plaque accumulation.}
\label{fig:qlf_images_classes}
\end{figure}

The QLF-images are a good example of images where learning invariant representation is crucial for good predictive performance. Typical examples of QLF-images for each of the three RF-PP classes are provided in Figure~\ref{fig:qlf_images_classes}.
As seen from this figure, these images were taken under various conditions such as slightly different focal distances, rotations, angles and not all images are perfectly centered or focussed to get better resolution. Besides ambient conditions during taking the pictures, the definition of every person is unique. Thus, there is a risk that standard models would overfit and learn variations in angles and distances which are not important for the plaque assessment.

\subsection{Influence of Multi-channel Representation}
For the experiments on the RF-PP plaque labels, the CNN model results in superior performance over the other classification models if all three colour channels were used. In the experiments on the RF-mQH and mSLP labels, an improvement was achieved when only the Green channel was added. Moreover, the standard deviation of the training performance tends to be narrower compared to when the model is applied on the Red channel only. This is especially true for GBC, LR and both of the SVMC models.

The Red over Green ratio of pixel values is generally used to identify red fluorescent plaque. Therefore, previous work performed on QLF-images \cite{lee2013association,kim2014monitoring} used the Red over Green pixel intensities' ratio instead of using the Red channel's pixel intensity values only. The Green channel helps to distinguish plaque from gingiva, since they have slightly different pixel values in Green channel of RGB representation. As for adding the Blue channel, due to technical implementation of the QLF-camera, the blue backscattered light is expected to produce sharper defined edges in images with little red fluorescent plaque, in comparison to images with a thicker plaque. The CNN model incorporates usage of all three colour channels without calculating ratios, thus numerically it is more stable and preferable. Based on these results we conclude that the CNN model benefits from multi-channel representation of the images. Precisely speaking, the CNN model efficiently and explicitly uses the fact that each colour channel contains important information relevant to the classification task.

\section{Conclusion}
In this study, we applied the CNN model for the automatic classification of red fluorescent dental plaque images. A comparison with several other state of the art shallow classification methods clearly showed the advantage of the CNN model in achieving a higher prediction performance. Such a result was possible because the CNN model directly learns invariant feature representations from raw pixel intensity values without engineering of hand-crafted features. We expect that Deep Learning of red fluorescent dental plaque images can help dental practitioners to perform efficient fluorescent plaque assessments and thus contribute to the improvement of patients' oral health.


\end{document}